\documentclass[graybox]{styles/svmult}

\usepackage{mathptmx}       
\usepackage{helvet}         
\usepackage{courier}        
\usepackage{type1cm}        
\usepackage{makeidx}         
\usepackage{graphicx}        
\usepackage{multicol}        
\usepackage[bottom]{footmisc}

\AtBeginDocument{%
  \providecommand\BibTeX{{%
    \normalfont B\kern-0.5em{\scshape i\kern-0.25em b}\kern-0.8em\TeX}}}

\usepackage [english]{babel}
\usepackage [autostyle, english = american]{csquotes}
\MakeOuterQuote{"}
\usepackage{latexsym}
\usepackage{color}
\usepackage{colortbl}
\usepackage[colorlinks]{}
\usepackage{pgfplots,wrapfig}
\usepackage{tikz}
\usepackage{booktabs}

\newcolumntype{P}[1]{>{\centering\arraybackslash}p{#1}}
\pgfplotsset{compat=1.12}
\usepackage{array, caption, floatrow, tabularx, makecell, booktabs}%
\usepackage{enumitem}

\usepackage{subcaption}
\usepackage{xcolor}
\usepackage{soul}
\usepackage{booktabs}

\definecolor{lightyellow3}{HTML}{fff2cc}
\definecolor{tealcustom}{HTML}{afded9}
\definecolor{redcustom}{HTML}{deb4af}

\newcommand{\hlc}[2][yellow]{{%
    \colorlet{foo}{#1}%
    \sethlcolor{foo}\hl{#2}}%
}

\usepackage{url}

\usepackage{bold-extra}

\usepackage{tikz}
\usetikzlibrary{positioning,calc}

\tikzset{
  redondo/.style={
    draw=blue,
    line width=1pt,
    rounded corners=3pt,
    text width=#1
  },
  punto/.style={
    fill=red,
    circle,
    inner sep=1.25pt
  },
  tresp/.pic={
    \node[punto] at (0.25,0) {};
    \node[punto] at (0.5,0) {};
    \node[punto] at (0.75,0) {};
  },
  dosp/.pic={
    \node[punto] at (0.25,0) {};
    \node[punto] at (0.5,0) {};
  },
  cuadra/.style={
    fill=teal,
    minimum size=10pt
  },
  arr/.style={
    line width=1pt,
    draw=green!70!black,
    ->,
    >=latex
  }  
}

\makeindex             

\begin{document}

\title*{Let's Get Personal: Personal Questions Improve SocialBot Performance in the Alexa Prize }
\titlerunning{Personal Questions Improve SocialBot Performance in the Alexa Prize}

\author{Kevin K. Bowden and Marilyn Walker}
\institute{Kevin K. Bowden and Marilyn Walker \\
Natural Language and Dialogue Systems Lab, University of California, Santa Cruz \\ 
\email{{kkbowden, mawalker}@ucsc.edu}}

%
%
\maketitle

\abstract{There has been an increased focus on creating conversational open-domain dialogue systems in the spoken dialogue community. Unlike traditional dialogue systems, these conversational systems cannot assume any specific information need or domain restrictions, i.e., the only inherent goal is to converse with the user on an unknown set of topics. While massive improvements in Natural Language Understanding (NLU) and the growth of available knowledge resources can partially support a robust conversation, these conversations generally lack the rapport between two humans that know each other. We developed a robust open-domain conversational system, Athena, that real Amazon Echo users access and evaluate at scale in the context of the Alexa Prize competition. We experiment with methods intended to increase intimacy between Athena and the user by heuristically developing a rule-based user model that personalizes both the current and subsequent conversations and evaluating specific personal opinion question strategies in A/B studies. Our results show a statistically significant positive impact on perceived conversation quality and length when employing these strategies.}

\section{Introduction}
There has recently been an increased focus on creating dialogue systems capable of open-domain social dialogue~\cite{konrad2021alquist,finch2020emora,gunrock2018,fang2018sounding}. Unlike traditional systems, these conversational dialogue systems cannot assume domain restrictions or any specific information need, i.e., the only inherent goal is to converse with the user. While modern systems have access to more information and better tools, foundational components of natural human-human conversation remain elusive, i.e., intimacy and agency. Combined, these should increase engagement in the conversation.

Our conversational open-domain dialogue system, Athena~\cite{harrison2020athena,walker-etal-2021-athena}, 
 is a competitor in the Amazon Alexa Prize ~\cite{gabriel2020further,hu2021further}, which gives real Amazon Echo users the ability to anonymously access and evaluate open-domain dialogue systems. These systems must interact socially, about any topic, with the goal of achieving a highly rated (\textgreater 4 out of 5 average user rating) conversation that lasts longer than 20 minutes. Further detail on  Athena's technical components are available in the associated technical reports~\cite{harrison2020athena,walker-etal-2021-athena}.

In general, there are three primary challenges with open-domain conversation:

\begin{enumerate}
    \item[P1:] There are an infinite number of valid topics we must support. While some topics are common and have been explored in previous dialogue research, e.g., Movies and Music, other topics reflect a user's special interest. Moreover, since Athena is a live system,  there is an implicit expectation that Athena has up-to-the minute live data
    .
    \item[P2:] We cannot rely on the user having an explicit goal or information need; they may just want to chat for entertainment. 
    \item[P3:] The conversation must be social. This means that we cannot artificially bloat the length of the conversation by telling twenty jokes  in a row. Instead we must be responsive to a user's interest in  different topics. 
\end{enumerate}

In this paper we focus on personalization as a partial solution for all three of these core problems. We cannot remove topics from the pool of valid topics in \textbf{P1}, however by learning more about the user's interests we can direct the  conversation to areas of Athena's knowledge. To accomplish this, we can start the  conversation with personal questions that help Athena learn about the user. Whether or not the user has an explicit goal in mind for \textbf{P2}, we can share control of the conversation by using personal opinion questions to engage the user on a level beyond pure fact retrieval. This process will help direct users to content about which Athena has opinions. Personal questions are also a crucial part of human-human social conversation, as in \textbf{P3}. Most human-human conversations would quickly end if one of the conversational partners feel their interests are ignored. It's vital, then, that Athena displays high levels of responsiveness to demonstrate her listening abilities and show understanding~\cite{reis2011familiarity, reis1996attachment}. Current research has also found in dyadic conversations that a conversational partner was more well liked if they ask more follow-up questions~\cite{huang2017doesnthurttoask}: a similar strategy could increase the rapport between the user and Athena.


Existing work also shows the impact of experience with dialogue systems when engaging with a system like Athena~\cite{bowden2019entertaining}. Besides varying levels of experience, there are several  user pools with different needs: there are users who have never interacted with Athena vs. repeat users, kids vs. adults, and users with varied backgrounds and interests. The only way to effectively cater to all of them is to personalize each conversation for each user. For example, the personal opinion questions we discuss below  are labeled with age appropriateness, e.g., when talking about Movies with a child, Athena talks about cartoons like Frozen. Participation in the Alexa Prize allows us to evaluate our hypotheses:  asking  questions at the start of the conversation allows Athena to populate a personalized user model, users are willing to give highly informative feedback if invited, and A/B testing with live traffic shows a statistically significant positive impact on both perceived conversation quality and length when using personal opinion questions. 

The remainder of this paper is organized as follows: Section~\ref{sec:rw} examines the current state of open-domain conversational systems with a specific emphasis on the Amazon Alexa Prize. Section~\ref{sec:user_pref_and_interests} demonstrates Athena's strategy of front-loading the conversation with personal questions to quickly build a personalized user model. Subsequently,  Section~\ref{section:opinion_interests_hobbies_evaluation} analyzes the resultant trends established across a large pool of real Alexa users while Section~\ref{section:user_advice_and_questions} explores the utility of asking the users directly for feedback and inviting personal questions from users. Then, in Section~\ref{section:wyr} we investigate two specific personal opinion question strategies (asking topic related Would You Rather choices and open-ended Hypothetical questions), for which we see a statistically significant improvement in both rating and length when evaluated at scale. In Section~\ref{section:conc} we conclude. 


\section{Related Work}
\label{sec:rw}

Dialogue systems have been an area of interest for over 50 years~\cite{weizenbaum1966eliza}. Much previous work with dialogue systems has been centered around the explicit goal of completing a task, e.g.,~booking a flight, or booking a restaurant. Commonly, these systems anticipate an explicit goal or "information need"~\cite{kiseleva2016predicting,chuklin2015click,Radlinski17}. Our goal of making a casual social dialogue system that is  open domain is a very distinct task. 

There have been many different approaches to open-domain dialogue~\cite{ma2021unstructured}. Recent work has seen an increased emphasis on trained end-to-end conversational systems~\cite{sordoni2015neural,vinyals2015neural,mila2017,burtsev2018first}. However, we believe that  these models are not yet ready for real user interactions~\cite{adiwardana2020meena,zhang2020dialogpt,roller2021recipes}. Instead, many Alexa Prize systems take a hybrid approach of combining rules, retrieval, and generation to  create a more robust system~\cite{song2016twoAreBetterThanOne,fedorenko2018avoiding,zhou2020xiolice}. These hybrid approaches closely resemble Athena's design.

Other work that is focused on open-domain personalization has been focused on short exchanges~\cite{yang2017personalizedResponseGenViaDomainAdaptation}, or used   Twitter  and TV Script data as a source of utterances~\cite{li2016persona}. This is not adequate for Alexa Prize conversations~\cite{bowden2018opendomain}. 

Athena's unique environment, the Amazon Alexa Prize~\cite{chandra2018conversational,gabriel2020further}, requires open-domain social chitchat with a spoken dialogue system. Over the last four years several competitive systems have emerged~\cite{konrad2021alquist,finch2021emora,liang2020gunrock,fang2018sounding}. Personalization has become an increasingly important part of these social systems. IrisBot, Genuine2, and Caspr use repeat user detection to slightly alter some responses~\cite{ahmadvand2018iris,kinjal2021caspr,rodriguezCantelar2021genuine}. 
SoundingBoard associated their users with Big-5 personality traits and adapted the conversation to those traits ~\cite{fang2018sounding}. Tartan and Emora personalized conversation by using user-demographic information with  handcrafted rules to affect follow-up content and ask  an increased number of personal questions~\cite{chen2020tartan,finch2021emora}. Alana used user profiles to track the interests of returning users and asked personal questions to build rapport~\cite{alana2018}. Gunrock maintained user profiles which were used to both affect follow-up content and change sub-topic discussions according to  gender, i.e., Fashion~\cite{liang2020gunrock}. Audrey trained a reinforcement learning model tasked with personalizing topic switching~\cite{hong2020audrey}. Proto uses an unspecified slot extraction process during the natural language understanding phase to extract attributes later used when personalizing responses~\cite{saha2021proto}, while Dream used Reddit data with the BART language model~\cite{lewis2020bart} to create a topic-based vectorization of Reddit user personas, which was subsequently used to give topic recommendation to real users~\cite{baymurzina2021Dream}. 
Alquist experimented with personalized sub-dialogue selection~\cite{konrad2021alquist}, while Athena reported results using a personalized topic promotion strategy~\cite{walker-etal-2021-athena}. This work is closely related to our own; however none of these systems evaluate and report the direct results of asking personal opinion questions, nor do they present the trends captured by their user models/personas.

We targeted two playful personal opinion question strategies (Would You Rather and Hypothetical questions) because playfulness is considered important when building rapport between conversational agents and users~\cite{shani2022alexasnowman} and humor has been shown to increase likeability and trust~\cite{meyer2015understanding}. Would You Rather questions have also been used in previous therapeutic settings to encourage conversation~\cite{fields2009would}.

\section{Gathering User Preferences, Interests, and Advice}
\label{sec:user_pref_and_interests}

\begin{figure*}[h!t]
    \begin{tabular}{p{.35in}|p{4.1in}} 
    \toprule
        {\bf [A6]} & Cool, what have you been out doing?\\
        {\bf \cellcolor[HTML]{fff2cc}[U6]} & {\cellcolor[HTML]{fff2cc} \textbf{swim}}\\
        {\bf [A7]} & swimming? I'm glad to hear that you've been able to engage in your interests. It seems like the world has completely changed over the last year. Have you been able to work from home?\\
        {\bf \cellcolor[HTML]{fff2cc}[U7]} & {\cellcolor[HTML]{fff2cc} \textbf{i don't work but i've been able to do school}}\\
        {\bf [A8]} & Well, I hope you're able to keep up with your school work remotely. Does it get hard at times?\\
        {\bf \cellcolor[HTML]{d9d2e9}[U8]} & {\cellcolor[HTML]{d9d2e9} not really}\\
        {\bf [A9]} & Hm. Well, thinking about it, I kind of have to stay at home all the time too. Hey, I know it's probably not possible right now, but I've always wanted to go on a vacation and to travel. What's a place that you've always wanted to travel to?\\
        {\bf \cellcolor[HTML]{fff2cc}[U9]} & {\cellcolor[HTML]{fff2cc} \textbf{hawaii}}\\
        {\bf [A10]} & I've heard that it's very nice there. For me, I think the great wall of china is at the top of my list of travel destinations. It seems like it would be amazing to walk along it. So, Is there a particular reason why you want to go to hawaii?\\
        {\bf \cellcolor[HTML]{d9d2e9}[U10]} & {\cellcolor[HTML]{d9d2e9} i've already been there and i really liked it}\\
        {\bf [A11]} & Nice! Hey, I'm curious, What do you like the most about going somewhere?\\
        {\bf \cellcolor[HTML]{d9d2e9}[U11]} & {\cellcolor[HTML]{d9d2e9} just bring out with to feel like i don't have any responsibility there because it's not my own house}\\
        {\bf [A12]} & That's an interesting answer. I really like the culture! For example, I would love to see all the paintings in the Louvre. Hmm, Did you go on family vacations as a kid?\\
        {\bf \cellcolor[HTML]{d9d2e9}[U12]} & {\cellcolor[HTML]{d9d2e9} yeah}\\
        {\bf [A12]} & That's cool. While I would love to go on vacation, that isn't really an option at the moment. But What do you normally like to do for fun.\\
        {\bf \cellcolor[HTML]{fff2cc}[U13]} & {\cellcolor[HTML]{fff2cc} i \textbf{play chess}}\\
        {\bf [A13]} & Oh, I see. chess sounds like a really interesting weekend activity! I wish I had more of a body, so that I was able to do some fun activities on the weekend. ...
        Hey, is it alright if I ask for a little advice? Do you have any ideas how I could be more interesting?\\
        {\bf \cellcolor[HTML]{d9d2e9}[U15]} & {\cellcolor[HTML]{d9d2e9} if you had a different voice all the time }\\
        {\bf [A15]} & Thanks for the advice! You know, I realize I've been asking you a lot of questions, but how about me? Is there anything you want to know about me?\\
        {\bf \cellcolor[HTML]{d9d2e9}[U16]} & {\cellcolor[HTML]{d9d2e9} how old are you}\\
        {\bf [A16]} & I like to think of myself as being 20 something.
        \\
        \bottomrule
    \end{tabular}
    \caption{Example of Athena's introductory dialogue strategies. \textbf{A} and \textbf{U} represent turns from Athena and the user respectively. Turns that update the user model are highlighted in yellow.  \label{table:introduction_um_example}}
\end{figure*} 

Similar to when two humans are first meeting, it is challenging for Athena to break the ice with a new user; Athena has no idea where to direct the conversation in-order to engage the user. In this paper we argue that personalization increases the sense of intimacy and engagement. Therefore, Athena aims to build a user model as quickly as possible. To this end, Athena starts each new conversation with a carefully crafted introduction sequence that is explicitly designed to elicit user interests. Athena employs several different dialogue strategies to elicit this information. We ask the user for their name and learn about how the pandemic has impacted them. We also ask them about their hobbies, weekend activities, and vacation plans before inviting them to learn more about Athena by soliciting advice and inviting questions. By front-loading the conversation with these probing questions, we increase the chances of the user model picking up usable information. 

In Figure~\ref{table:introduction_um_example} we provide an excerpt of a conversation and introductory sequence. During this conversation, the user model learns the user's name,  that their hobbies include swimming and chess (\textbf{U6} and \textbf{U13}), that they are a student (\textbf{U7}), and that Hawaii is a travel destination of interest (\textbf{U9}). Some of this content will be immediately useful in the conversation, e.g., reaffirming their name and asking why they like Hawaii, while other pieces of information will be useful when initiating a new topic, e.g., talking about Sports because the user is a swimmer, or Board Games because of their interest in chess.

\newpage
Athena supports 17 topics ranging from large general topics, e.g., Sports, to smaller niche topics. Each topic is associated with  a set of referential expressions, for example, baseball, football, and Stephen Curry, all refer to the Sports topic. We also curated a gazetteer of $\sim$250 hobbies by analyzing the results of probing questions over several months. Each hobby is annotated with paraphrases, e.g., \emph{i like to paint}, \emph{i like painting}, and \emph{i painted when i was young} will all match the \emph{painting} hobby, and any associated topics when relevant, e.g., basketball and piano map to Sports and Music respectively. 

\subsection{Conversational Data Analysis}
\label{section:opinion_interests_hobbies_evaluation}

A key component when building a personalized user model is tracking hobbies and interests. Identifying common interests help practitioners select desirable topics in future conversational systems. Meanwhile, the knowledge Athena gains personalizes topic management which improves conversations~\cite{walker-etal-2021-athena}. By navigating to topics that the user likes, Athena is also able to ask additional personal questions later in the conversation that are more likely to interest the user (Section~\ref{section:wyr}). The data presented was collected by conversing with anonymous users during the Alexa Prize competition over several weeks, i.e., several thousand conversations with a majority of users talking to Athena for the first time. 

\begin{figure}[h!t]
    \centering
    \includegraphics[width=\linewidth]{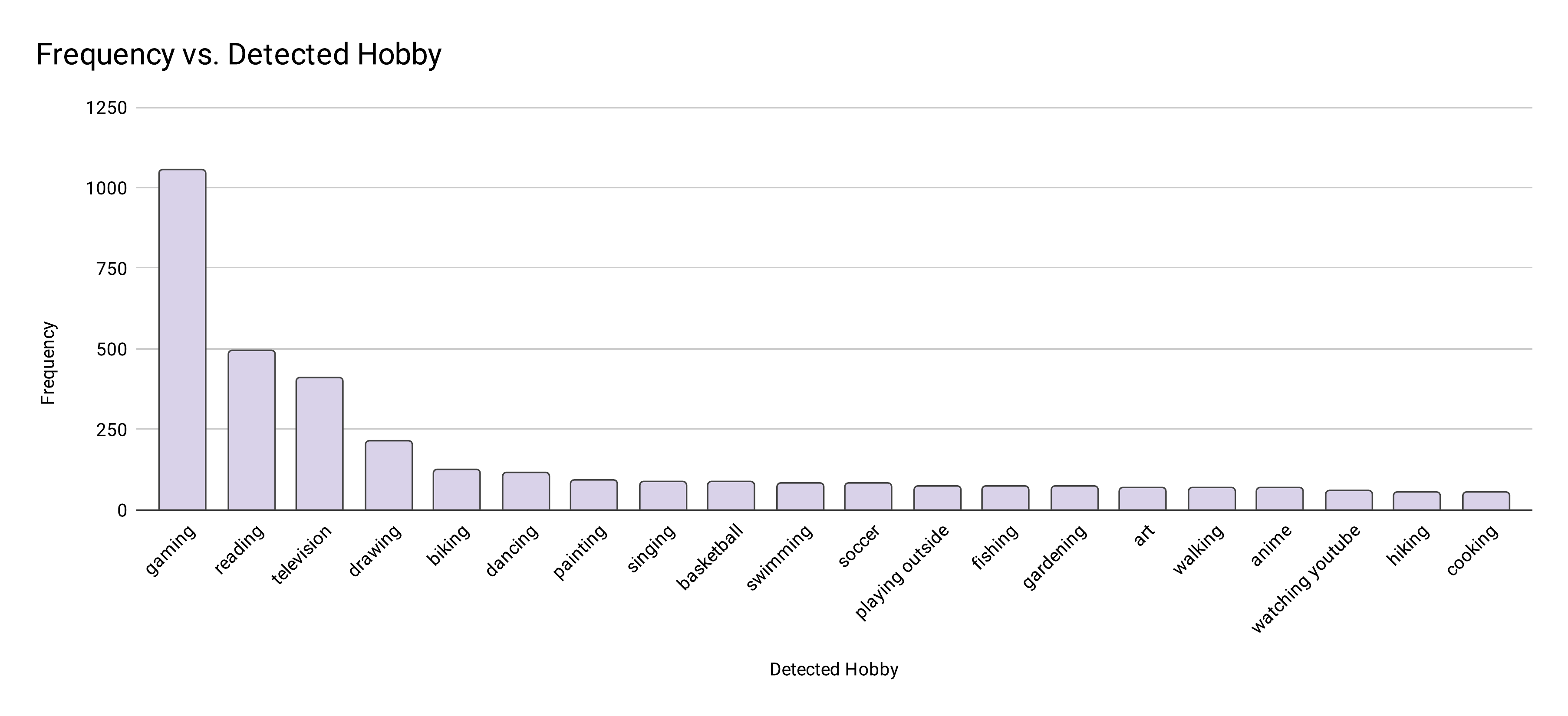}
     \vspace{-.3in}
    \caption{The distribution of the top 20 detected hobbies over a 1 month period.}
    \label{fig:detected_hobbies}
\end{figure}

Figure~\ref{fig:detected_hobbies} shows the distribution of the top 20 hobbies detected by the user model over a 32 day period of time. Only 42 out of 131 total detected hobbies were associated with more than 25 unique users. The top five hobbies, i.e., (video) gaming, reading, television, drawing, and biking, represent 48\% of all detected hobbies over this period of time. Gaming in particular, represents 22\% of all detected hobbies (1056 out of 4790).

\begin{figure}[h!tb]
    \centering
    \includegraphics[width=\linewidth]{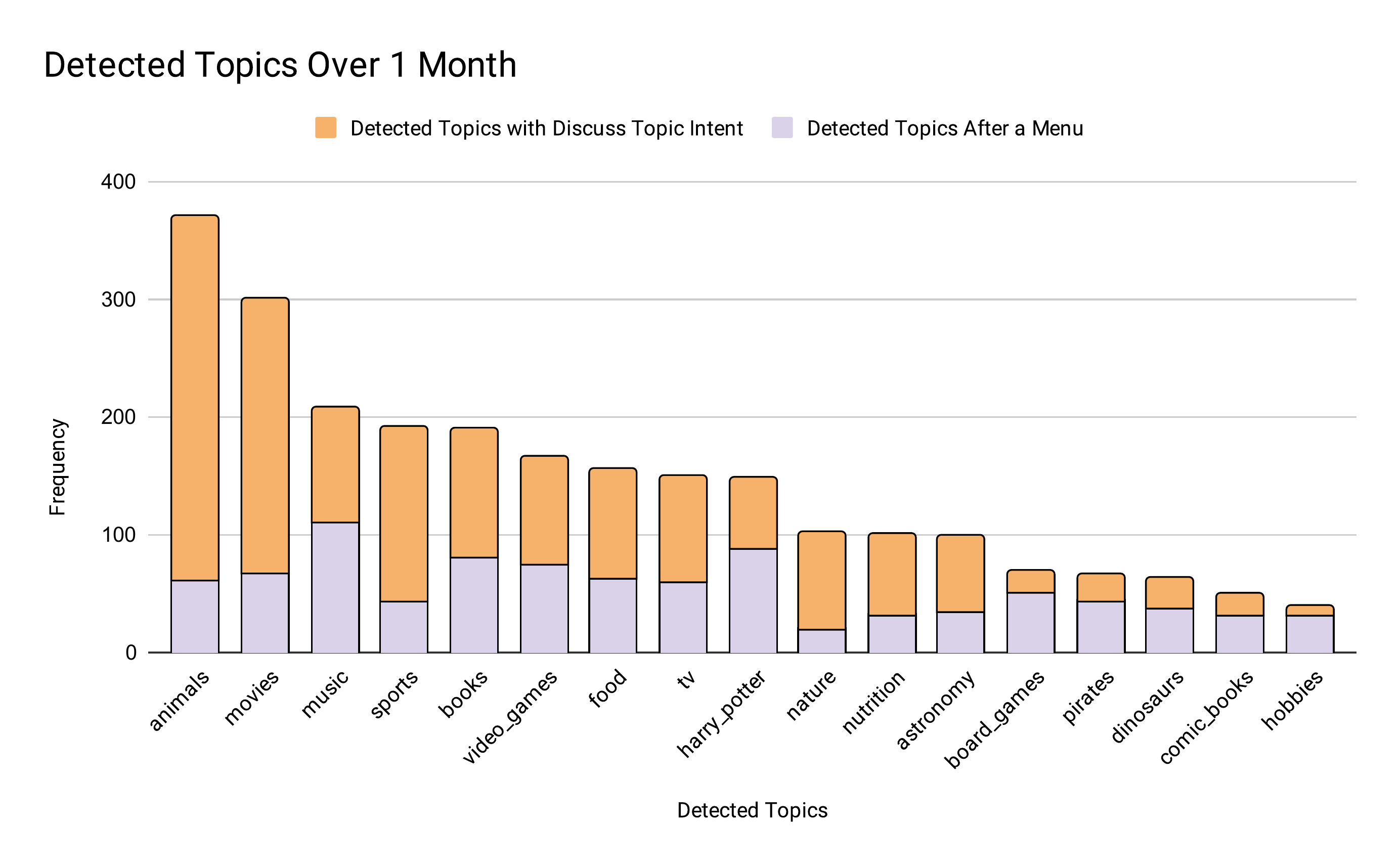}
    \vspace{-.3in}
    \caption{The frequency of a detected topic with an explicit discuss topic marker (orange) and the frequency of a detected topic after an explicit menu of choices (purple) over a 1 month period. }
    \label{fig:both_detected_topic_frequencies}

\end{figure}

Figure~\ref{fig:both_detected_topic_frequencies} provides the number of explicit discuss topic requests per topic, e.g., out of the blue the user says \textit{let's talk about movies}, and the number of explicit topic requests following a menu of options provided by Athena, e.g., the user says something like \textit{let's talk about music} after Athena prompts them with \textit{are you interested in talking about animals, books, or music?}. Figure~\ref{fig:both_detected_topic_frequencies} helps us understand which topics are most highly trafficked. Animals and Movies are the most frequent user-initiated topics (orange), likely because Movies is a commonly discussed topic in the Alexa Prize, and Animals is triggered by a user mentioning their pet. Meanwhile, Music and Harry Potter are the most common topic choice after the user has been given a menu of choices. This could indicate that users are interested in these topics, but do not necessarily expect that a socialbot will be able to discuss them. This trend is seen in other niche Athena topics, e.g., Comic Books and Dinosaurs. 

\begin{figure}[h!tb]
    \centering
    \includegraphics[width=\linewidth]{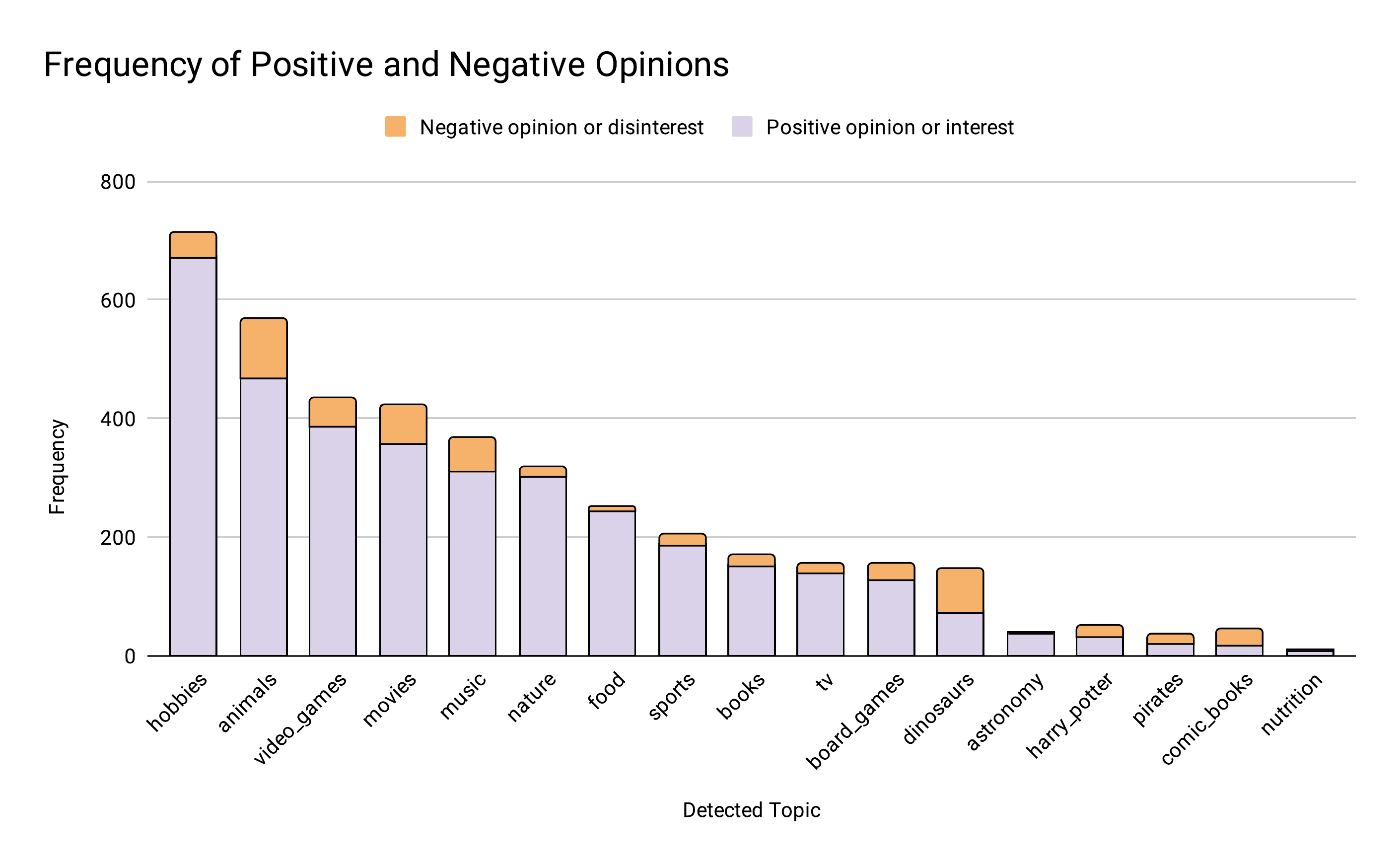}
     \vspace{-.3in}
    \caption{The frequency of each detected topic in a positive opinion/explicit interest, or negative opinion/explicit disinterest.}
    \label{fig:detected_topic_opinions}
    
\end{figure}

Over a 22 day period, we collected 11,415 opinions from 2,521 conversations. Users more commonly provided positive opinions than negative opinions (9,755 positive vs. 1,866 negative). Figure~\ref{fig:detected_topic_opinions} shows the distribution of detected topics in positive and negative opinions. The trends are reasonable; people like talking about their hobbies and their pets, but may not have an opinion about niche topics. 




\subsection{User Advice and Questions}
\label{section:user_advice_and_questions}

Athena further breaks the ice by asking the user one of three possible open-ended questions related to Athena's self-improvement (e.g., \textbf{A13} in Figure~\ref{table:introduction_um_example}) and to learn which topics the user finds interesting. The question we ask is always prefaced by the statement: \textit{Hey, is it alright if I ask for a little advice?}. The following represents Athena's three open-ended questions:

\begin{itemize}
    \item[\textbf{IceQ1}] Do you have any ideas how I could be more interesting?
    \item[\textbf{IceQ2}] I'm trying to figure out fun things to talk about. Would you mind telling me what kind of topics you like talking about with your friends?
    \item[\textbf{IceQ3}] I'm trying to figure out fun things to talk about. What are your personal interests and favorite conversational topics?
\end{itemize}

We collected $\sim$2,300 responses to these questions. While some users refused to participate or even became adversarial during this sequence, many others answered Athena's call with genuine user feedback. From these responses we can estimate topics of interest to each specific user, and for future system improvements. 

Responses to Athena's first question, \textbf{IceQ1}, are particularly informative; a sampling of these responses can be seen in~Figure~\ref{table:direct_user_feedback}. Some feedback is immediately useful for system development, such as reducing response latency and tweaking the way Athena talks through Speech Synthesis Markup Language (SSML), while other feedback helps us understand other desired functionalities of Athena, e.g., supporting commonly requested topics and "spirited debates". Users are also directly stating they want a more "personable" experience where Athena asks "personal questions", engages in topics of mutual interest, and discusses everyday events in their life such as work or school (as in our introductory excerpt in Figure~\ref{table:introduction_um_example}). 

\begin{figure}[t]
    \centering
    \includegraphics[width=\linewidth]{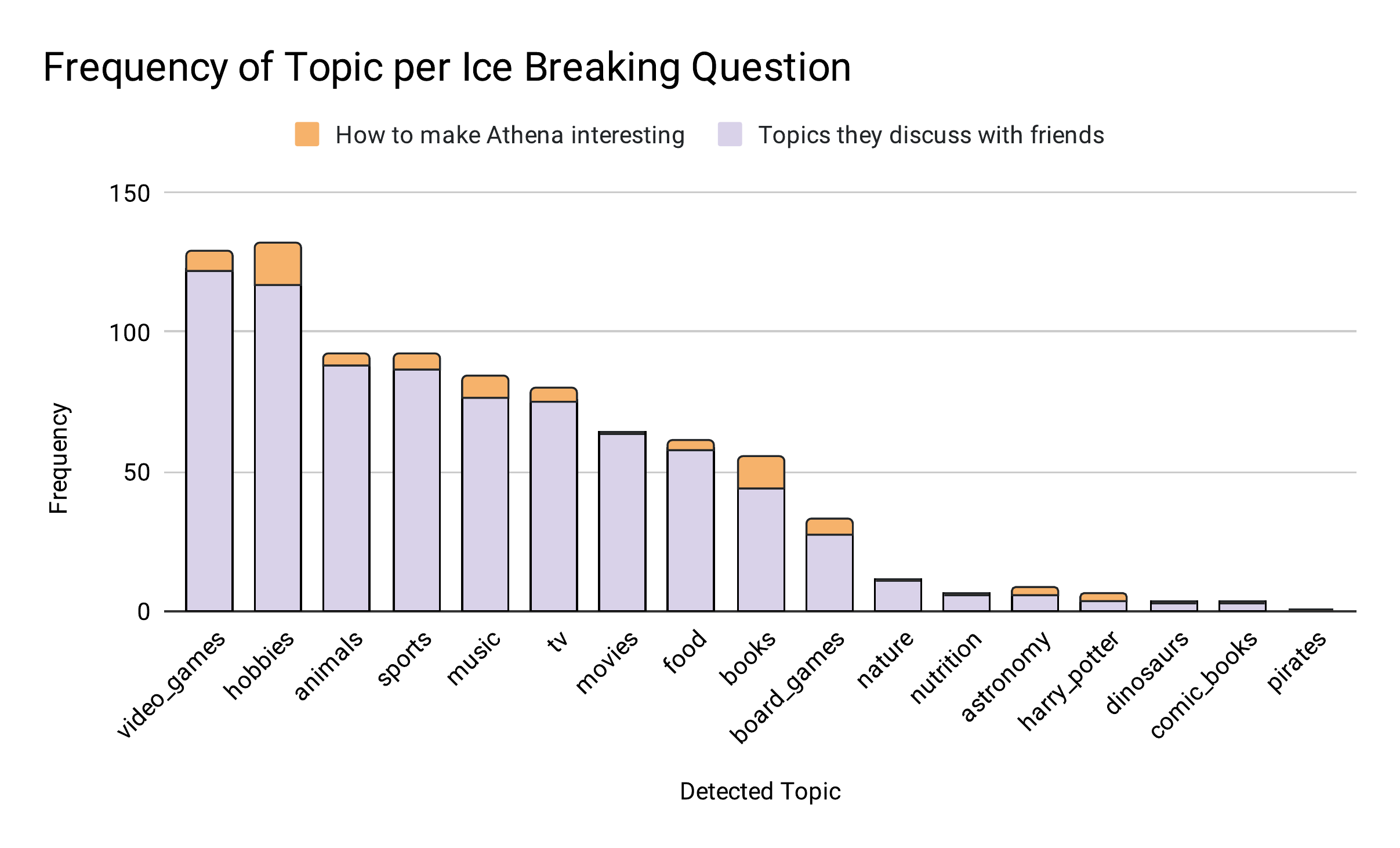}
    \vspace{-.3in}
    \caption{The frequency of a detected topic in response to ice breaking questions. }
    \label{fig:detected_topic_ice_breakers}
\end{figure}

\textbf{IceQ2} and \textbf{IceQ3} both share a similar goal; we want the user to explicitly tell us the types of things they like talking about. The feedback related to specific hobbies (e.g., gardening) and topics (e.g., animals and hobbies) that Athena can identify will be leveraged throughout the conversation when Athena is initiating new topics. 
In Figure~\ref{fig:detected_topic_ice_breakers} we see the distribution of topics detected across the $\sim$2300 responses. \textbf{IceQ1} is differentiated from \textbf{IceQ2} and \textbf{IceQ3}, since the questions have different intentions. Summing these results, we find that 859 topics were identified in 691 responses (some responses provided more than one topic). This means that asking a single ice breaking question at the start of a conversation results in personalizable data for a system supporting similar topics to Athena $\sim$30\% of the time. 

An additional part of the ice breaking process is to explicitly invite the user to ask us a question, i.e., Athena states: \textit{You know, I realize I've been asking you a lot of questions, but how about me? Is there anything you want to know about me?} (as in \textbf{A15} from Figure~\ref{table:introduction_um_example}).  Again, some users  were adversarial, but several users did ask questions, primarily about learning personal information about Athena. Ignoring bad questions, e.g., antagonistic questions or ASR errors, over 90\% of questions asked were personal questions about Athena's life and opinions. Figure~\ref{table:direct_user_feedback} gives example user questions that were a result of Athena's explicit invitation. 

\begin{figure}[h!t]
    \makebox[\textwidth][c]{
    \begin{tabular}{c || c}
    \toprule
        User Feedback & User Questions \\
        \hline
        can we ask questions and you answer them & what's next after the alexa bot competition\\
        i would like to develop emotional connections & what is your favorite book\\
        be more funnier & how old are you\\
        maybe you can tell more about yourself & do robots and stuff like you have a birthday\\
        be more personable & what is your favorite food and color\\
        by asking people how was your day & do you have any friends\\
        maybe ask some more personal questions & do you have any pets\\
        ask me personal questions & what's your favorite video game\\
        well you can have a hobby & what do you do in your free time\\
        ask about people's personal life instead of general questions & do you listen to music\\
        i want you to ask about school or work & what's your name\\
        well you could ask people really weird questions & do you ever get lonely\\
        good to talk about peoples families and the things they love & oh yeah what's your favorite superhero\\
        we can talk about our favorite colors and more about us & would you rather be a strawberry or a cantaloupe\\
        you could learn what other people like and share with them & if you could be any animal which would you be\\
        maybe ask like more out there questions & where do you see yourself in twenty to thirty years\\
        \bottomrule
    \end{tabular}}
    \caption{User feedback on how Athena can be more interesting and questions the user asked Athena when solicited. \vspace{-.3in} \label{table:direct_user_feedback}}
    \end{figure} 

\vspace{-.3in}

\section{Experiments with Personal Opinion Questions}
\label{section:wyr}

Inviting an exchange of opinions about the current topic of discussion creates a more intimate experience that increases user satisfaction. We have curated a dataset of topic-annotated personal opinion questions, which are inter-weaved throughout the conversation~\cite{bowden2019entertaining}. The personal opinion questions are split into two different strategies, Would You Rather choices (WYR) and open-ended Hypothetical questions (HYP). We picked these strategies because playfulness and humor are important when building rapport and trust~\cite{shani2022alexasnowman,meyer2015understanding}. These strategies have also been characterized as good conversation starters~\cite{fields2009would} and several lists of questions exist across the internet\footnote{E.g., https://conversationstartersworld.com contains several lists. The wouldyourather and hypotheticalsituation Reddit communities also have 316k and 102k subscribers respectively.}. These sequences can also be characterized as a game\footnote{E.g., https://psycatgames.com/app/would-you-rather/.}; gamifying our personal opinion question strategies slightly may have a desirable impact since "gaming" is a popular hobby among users, i.e., Figure~\ref{fig:detected_hobbies}. As we will discuss in Section~\ref{section:data_collection}, it is also straightforward to crowdsource the collection of question/answer pairs per topic, and the limited scope of answer choices, especially true for WYR, makes it easier to build NLU that can understand the user's responses. 

Figure~\ref{table:wyr_hyp_intro_example} demonstrates a sample conversation using these strategies. In this example, Athena initiates the Dinosaurs topic in \textbf{A1}, and proceeds to interweave 6 different dinosaur themed dialogue strategies
: \textbf{A1}-\textbf{A2} gives dinosaur facts, \textbf{A3}-\textbf{A4} talks about dinosaur preferences, \textbf{A4}-\textbf{A5} focuses on dinosaur movies, \textbf{A6}-\textbf{A7} goes through a dinosaur WYR sequence, \textbf{A7}-\textbf{A8} talks about why dinosaurs are extinct, and \textbf{A9}-\textbf{A10} goes through a HYP sequence. Athena's two personal opinion strategies extend topical depth by four turns. The user also provides us with long thoughtful responses, which indicate their engagement~\cite{walker2021IWSDSParadise}, furthering the sense of rapport built between Athena and the user.


\begin{figure*}[h!t]
    \begin{tabular}{p{.35in}|p{4.1in}} 
    \toprule
    {\bf [A1]} & But most people seem to find dinosaurs really interesting. There are dinosaur movies and whole floors of natural history museums dedicated to dinosaurs. Personally, I find it fascinating that modern birds are descended from a particular family of dinosaurs.\\
    {\bf \cellcolor[HTML]{d9d2e9}[U1]} & {\cellcolor[HTML]{d9d2e9} hella cool}\\ 
    {\bf [A2]} & Definitely.	I was surprised to hear how much work went into creating the movie Land Before Time. Over 600 background paintings were made for the film. The artists had to create a believable environment in which there was almost no foliage. Most paintings depicted beautiful but barren wastelands.\\
    {\bf \cellcolor[HTML]{d9d2e9}[U2]} & {\cellcolor[HTML]{d9d2e9} surprising}\\ 
    {\bf [A3]} & Right?	My favorite dinosaur is the Tyrannosaurus Rex. It's huge and ferocious, but the little arms make it somewhat adorable. What about you? What's your favorite dinosaur?\\
    {\bf \cellcolor[HTML]{d9d2e9}[U3]} & {\cellcolor[HTML]{d9d2e9} i'm not sure i have to say a t. rex}\\ 
    {\bf [A4]} & Okay. There are lots of movies about dinosaurs. One of the most famous dinosaur movies is Jurassic Park, old but classic. Are there any dinosaur movies that you love?\\
    {\bf \cellcolor[HTML]{d9d2e9}[U4]} & {\cellcolor[HTML]{d9d2e9} i have to say jurassic park to adjust because that's the only one i've seen}\\ 
    {\bf [A5]} & Awesome! That one is well known! I seem to remember people saying the dinosaurs were very vivid\\
    {\bf \cellcolor[HTML]{d9d2e9}[U5]} & {\cellcolor[HTML]{d9d2e9} i never knew that}\\ 
    {\bf [A6]} & yeah, I love to collect these fun facts! But anyways, I'm interested in your opinion. \hlc[tealcustom]{Would you rather ride on the back of a pterodactyl or brontosaurus as your primary means of transportation?}\\
    {\bf \cellcolor[HTML]{fff2cc}[U6]} & {\cellcolor[HTML]{fff2cc} \textbf{i have to say brontosaurus and i hate heights}}\\ 
    {\bf [A7]} & \hlc[tealcustom]{Choosing a Brontosaurus you say, great pick! I would rather ride a pterodactyl so I could experience flight as well as move around more efficiently. Riding a brontosaurus would be thrilling because of their size but less practical than a pterodactyl.} So, okay. Do you know why dinosaurs are extinct?\\
    {\bf \cellcolor[HTML]{d9d2e9}[U7]} & {\cellcolor[HTML]{d9d2e9} yeah cause i'm eat your head it right}\\ 
    {\bf [A8]} & Ah nice, you must be super smart. Let me tell you what I heard.The mass extinction of dinosaurs is a scientific mystery, but most hypotheses include an extraterrestrial impact or a massive bout of volcanism. What do you think happened?\\
    {\bf \cellcolor[HTML]{d9d2e9}[U8]} & {\cellcolor[HTML]{d9d2e9} i'd say}\\ 
    {\bf [A9]} & I'd like to hear your ideas on this. \hlc[tealcustom]{Let's say you are trapped on Jurassic World alone with the dinosaurs, what would you do to survive?}\\
    {\bf \cellcolor[HTML]{fff2cc}[U9]} & {\cellcolor[HTML]{fff2cc} \textbf{i'm not sure i just say find sheltering something and start there}}\\ 
    {\bf [A10]} & \hlc[tealcustom]{It's a hard question! I would try to become closer to the younger dinosaurs so that they would grow up with me and feel more comfortable with me. I would also try to help them by finding food and shelter for them.} At any rate, Dinosaurs... 
    \\
    \bottomrule
    \end{tabular}
    \caption{An example conversation about Dinosaurs with both of the personal opinion strategies, increasing topical depth. \textbf{A} and \textbf{U} represent turns from Athena and the user respectively. 
    \label{table:wyr_hyp_intro_example}}
\end{figure*}

\subsection{Data Collection}
\label{section:data_collection}

We collected a novel dataset in which we asked crowdsourced workers to provide both questions and answers for specific topics such as Food, Nature, and Astronomy~\cite{bowden2019entertaining}. 
After filtering out low-quality and extraneous topics there are 635 pairs across 14 topics. We also annotate our data for kid-friendliness, which yields 342 question/answer pairs spanning all 14 supported topics marked as kid-friendly\footnote{Data available from https://nlds.soe.ucsc.edu/software.}. 


When a response generator is deciding to use our personal opinion questions, it will first determine if the user is a child (over a 22 day period 11.44\%~(641) of users were identified as youths by our user model). 
While the age-appropriateness of some content may be subjective, Athena aims to err on the side of caution.

These questions were written and answered by human workers, which means the answer content inherently assumes human properties. However, users are quick to point out that, for example, Athena does not eat, go outside, or have a physical body, which can cause breakdowns mid-conversation. 
Therefore, we curated this dataset to ensure all of the system's responses were realistic for an Echo device. For example, with Food Athena consistently reminds the user that she can only \textit{eat electricity}.


\subsection{Methodology}

Personal opinion questions are used to increase topical depth. When a user is engaged with one of the 14 topics that has exchanges, one of each question type (WYR and HYP) can be asked per topic per conversation. After asking a question, we try to match the user's answer to an expected answer. If Athena can make a match, we can customize the next grounding statement, otherwise we'll fallback to a generic grounding statement. After acknowledging the user's answer, the system will provide its own opinion. Finally, Athena appends a short transitional phrase to the end of its answer before initiating a new sub-dialogue. In Figure~\ref{table:wyr_hyp_intro_example}
we can see longer conversation snippets that use these personal opinion questions to increase topical depth. The dialogue manager has a preference to interleave personal questions with other types of exchange. We saw this preference in Figure~\ref{table:wyr_hyp_intro_example}, where the WYR (\textbf{A6}) and HYP (\textbf{A9}) questions are separated by other Dinosaur content.

\begin{figure*}[h!tb]
    \begin{tabular}{p{.5in}p{.6in}p{.35in}|p{3.in}} 
    \toprule
        {\bf Comic Books} & {\bf WYR} & {\bf [A1]} &  What are your thoughts on this. If you wanted a mentor to help use your powers, would you rather choose tony stark, or doctor strange?\\
        \cellcolor[HTML]{d9d2e9} & \cellcolor[HTML]{d9d2e9} & {\bf \cellcolor[HTML]{d9d2e9}[U1]} & doctor strange {\cellcolor[HTML]{d9d2e9}  }\\
        \hline
        \hline
        {\bf Animals} & {\bf WYR} & {\bf [A2]} & What are your thoughts on this. Would you rather tame and own a lion or a gorilla? \\
        \cellcolor[HTML]{d9d2e9} & \cellcolor[HTML]{d9d2e9} &  {\bf \cellcolor[HTML]{d9d2e9}[U2]} & i love both {\cellcolor[HTML]{d9d2e9}  }\\
        \hline
        \hline
        \hline
        {\bf Board Games} & {\bf HYP} & {\bf [A3]} & What are your thoughts on this. If you had the power to become the world's best board game player, what board game would you choose at excelling at? Why?\\
        \cellcolor[HTML]{d9d2e9} & \cellcolor[HTML]{d9d2e9} & {\bf \cellcolor[HTML]{d9d2e9}[U3]} & i would choose every game because it's always fun to play against other people and have good games{\cellcolor[HTML]{d9d2e9}  }\\
        \hline
        \hline
        {\bf Food} & {\bf HYP} & {\bf [A4]} & I'm super curious to hear your thoughts. If you could make one type of food healthy and have zero calories, carbs, etc what would it be?\\
        \cellcolor[HTML]{d9d2e9} & \cellcolor[HTML]{d9d2e9} & {\bf \cellcolor[HTML]{d9d2e9}[U4]} & that's a hard question {\cellcolor[HTML]{d9d2e9}  }\\
        \bottomrule
    \end{tabular}
    \caption{Example question/answer pairs. \textbf{A1}-\textbf{A2} represent a Would You Rather sequence, while \textbf{A3}-\textbf{A4} represent the more open-ended Hypothetical sequence.\vspace{-.15in} \label{table:wyr_and_hq_examples}}
    
\end{figure*}

In Figure~\ref{table:wyr_and_hq_examples}, we have further examples of user responses. Responses vary between the two question strategies. In WYR sequences (\textbf{U1}-\textbf{U2}), the user tends to provide short but precise responses, e.g., \textit{doctor strange} in \textbf{U1}, which often contain the exact text associated with an answer choice. In some cases, the user may answer without using the exact same text, or otherwise imprecisely as in \textbf{U2}. 
In HYP sequences (\textbf{U3}-\textbf{U4}), there are different answer trends. For example, some users embrace the open-ended nature of the questions by providing detailed responses, as in \textbf{U3} while other users might provide precise responses, e.g., \textit{risk} in an alternate response to \textbf{A3}, however, without the integration of extensive external knowledge resources, it is hard to determine the suitability of the answer. Meanwhile, other users may struggle to think of answer to an open-ended question on the fly, as in \textbf{U4}.

\subsection{Evaluation}
\label{section:wyr_hyp_eval}

\begin{table*}[h!t]
    \begin{tabular}{c|cc|ccc|ccc} 
    \toprule
        Req. WYR & A convs. & B convs. & A rating & B rating & p-value & A length & B length & p-value\\
        \hline
        0 & 763 & 232 & 3.71 & 3.78 & .709 & 22.49 & 22.05 & .774\\				
        1 & 313 & 232 & 3.94 & 3.78 & .125 & 37.39 & 22.05 & .000\\		
        2 & 111 & 232 & 4.27 & 3.78 & .000 & 58.26 & 22.05 & .000\\				
        3 & 51 & 232 & 4.38 & 3.78 & .002 & 71.77 & 22.05 & .000\\
        \bottomrule
    \end{tabular}
    \caption{Results from an A/B trial over 5 days. A represents a system with WYR enabled, while B represents the system with neither personal opinion question (POQ) enabled. The Req. WYR column represents the minimum number of WYR sequences in the conversation. \label{table:wyr_ab_results}}
\end{table*}  

\begin{table*}[h!t]
    \begin{tabular}{c|cc|ccc|ccc} 
    \toprule
        Req. HYP & A convs. & B convs. & A rating & B rating & p-value & A length & B length & p-value\\
        \hline
        0 & 1980 & 681 & 3.70 & 3.72 & .734 & 21.92 & 21.02 & .303\\
        1 & 804 & 681 & 3.84 & 3.72 & .085 & 35.98 & 21.02 & .000\\		
        2 & 282 & 681 & 3.92 & 3.72 & .032 & 53.84 & 21.02 & .000\\				
        3 & 104 & 681 & 4.03 & 3.72 & .034 & 75.14 & 21.02 & .000\\
        \bottomrule
    \end{tabular}
    \caption{Results from an A/B trial over 14 days. A represents a system with HYP enabled, while B represents the system with neither personal opinion question (POQ) enabled. The Req. HYP column represents the minimum number of HYP sequences in the conversation.\label{table:hyp_ab_results}}
\end{table*}

We ran two A/B experiments with live traffic, i.e., any person with access to Alexa is a possible participant. In the A case (75\% of traffic), just one type of the personal opinion question is enabled, while in the B case (25\% of traffic) neither is enabled. The main evaluation criteria for the Alexa Prize is to create long and engaging conversations. Therefore, we evaluate the two dialogue strategies with respect to two metrics: overall user rating and conversation length. User rating is direct feedback - after the conversation ends, the user rates the system on a scale from 1-5 based on how interested they would be in talking to our system again. Length is evaluated automatically based on the number of exchanges in the conversation. In Table~\ref{table:wyr_ab_results} and Table~\ref{table:hyp_ab_results} we can see the Would You Rather (WYR) and Hypothetical (HYP) results respectively. In both cases, we only consider conversations that lasted longer than 6 exchanges, to account for early or accidental hang-ups at the start of the conversation, which can negatively bias results~\cite{walker2021IWSDSParadise}.

Our WYR evaluation ran over 5 days while HYP ran for 14 days. We initially planned to run both evaluations for 14 days, however, we felt we had a sufficient sample size with WYR after only 5 days. In both evaluations, we systematically vary a threshold for the minimum number of instances of personal opinion questions (POQs) in the conversation. Since the user rating only comes at the end of the conversation, varying this threshold should make it easier to observe the impact of our variables on the conversation. When the minimum Req. WYR/HYP value is 0, all conversations longer than six exchanges are included. Each POQ sequence includes two exchanges - one to ask the question and another to answer. 

User ratings trend towards an improved experience when at last one POQ strategy is enabled, and becomes statistically significant once the threshold requires at least two POQ per conversation. When evaluating the Pearson correlational between the user ratings and the number of POQs, we see a weak, but statistically significant (p$<$.001), correlation: .17 and .10 for WYR and HYP respectively. WYR results in slightly higher ratings than HYP, and we can see the correlation between HYP and the user ratings is also weaker than WYR. We speculate that the difference in rating may be related to the increased difficultly of NLU when acknowledging HYP answers. HYP questions are open-ended and designed to provoke innumerable valid answers, while WYR only has two valid answers. Therefore, it is easier for Athena to detect and signal an understanding of the user's answer to a WYR question than a HYP question. An example of this scenario is in Figure~\ref{table:wyr_hyp_intro_example}. In \textbf{A7} Athena repeats the user's choice of \textit{brontosaurus} in a WYR sequence, but when responding to a HYP sequence in \textbf{A10}, the complexity of the user's answer forces Athena to answer without an explicit signal that acknowledges the user. This may suggest the importance of signalling when responding to personal questions, but requires further investigation.  

While rating is direct user input, it only comes at the end of the conversation. Therefore, we also calculate how much of the conversation was part of a POQ sequence. On average, when 1 POQ is required, this translates to at least $\sim$5.5\% of the conversation's total exchanges being part of a POQ sequence. This increases to at least $\sim$7.2\% of exchanges and $\sim$8.2\% of exchanges when 2 POQ and 3 POQ are required, respectively. In other words, as our threshold increases, the required percentage of the conversation that is part of a POQ sequence also increases. Since both the rating and POQ contribution are increasing, we feel confident that the impact of Athena's POQ strategies is not vanishing as length increases.   

In both A/B tests the difference in conversational length becomes statistically significant when we require at least 1 personal opinion question. Additionally, both WYR and HYP see a strong Pearson correlation between the length of the conversation and the number of POQ: .82 and .80 respectively (p$<$.001). Since conversation length is a good predictor of conversation quality~\cite{walker2021IWSDSParadise}, we interpret these results as confirmation that our POQs foster a more engaging user experience. 

Over a 22 day period, 5,113 POQs were asked across the 14 supported topics. Of these 5,113 questions, 4,494 (about 88\%) were answered in a way that allowed the conversation to continue on-topic. This validates the effectiveness of the POQ strategies at extending topical depth. Two topics (Nature and Food) make up the majority of the terminal cases, primarily due to their design.  



\section{Conclusion}
\label{section:conc}
In this paper we proposed three fundamental challenges practitioners face when building open-domain dialogue systems: covering an intractable landscape of topics, users who aimlessly meander through the conversation, and the core requirement of sociability. We propose a partial solution to these three problems lies in personalizing the conversation to each user. We implement a system, Athena, and deploy it at scale in the Amazon Alexa Prize, yielding several interesting results. Firstly, we front-load the conversation with a sequence of probing personal questions to rapidly model the user. By qualifying several thousand real user conversations we're able to understand the types of topics users are interested in along with common user hobbies - informing future practitioners where to concentrate their efforts. Secondly, we ask the user for candid advice on how to improve Athena and invite the user to ask questions. The user input is surprising informative, helping to identify areas of improvement and future development. Moreover, the user input indicates a strong desire for a more personable experience as evidenced by direct user requests, e.g., \textit{be more personable}, and by the fact that most user questions ($>$ 90\%) were personal questions about Athena's life and opinions. This data further signals the possible benefits of asking the user candid questions during a conversation for future evaluations. Finally, we evaluated two personal opinion question strategies, Would You Rather choices and open-ended Hypothetical questions. A/B tests carried out over several days of live user traffic confirmed a statistically significant positive impact on both conversation rating and length when these questions were included in a conversation.

\bibliographystyle{styles/spmpsci.bst}
\bibliography{iwsds_bib.bib}

\end{document}